\pdfoutput=1

\documentclass[11pt]{article}

\usepackage[final]{acl}

\usepackage{times}
\usepackage{latexsym}

\usepackage{makecell}
\usepackage{graphicx}
\usepackage{multirow}
\usepackage{amsfonts,amssymb}
\usepackage{booktabs}

\usepackage[T1]{fontenc}

\usepackage[utf8]{inputenc}

\usepackage{microtype}
\usepackage{amsmath}
\usepackage{bm}
\usepackage{pifont}

\newcommand{\MODELNAME}{MCL}
\newcommand{\FULLMODELNAME}{multi-perspective course learning}

\title{Pre-training Language Model as a Multi-perspective Course Learner}

\author{
Beiduo Chen$^{\S\ddag}$\thanks{\ \ Contribution during internship at Microsoft.}, Shaohan Huang$^\ddag$\thanks{\hspace{1.5mm} Corresponding author.}, {\bf Zihan Zhang$^\ddag$, Wu Guo$^\S$, Zhenhua Ling$^\S$,} \\
{\bf Haizhen Huang$^\ddag$, Furu Wei$^\ddag$, Weiwei Deng$^\ddag$ and Qi Zhang$^\ddag$} \\
  $^\S$~National Engineering Research Center of Speech and Language Information Processing, \\
      University of Science and Technology of China, Hefei, China \\
  $^\ddag$~Microsoft Corporation, Beijing, China \\
{\texttt {beiduo@mail.ustc.edu.cn},} {\texttt { \{guowu,zhling\}@ustc.edu.cn},} \\
{\texttt {\{shaohanh,zihzha,hhuang,fuwei,dedeng,qizhang\}@microsoft.com}}
}

\begin{document}
\maketitle
\begin{abstract}
ELECTRA~\cite{electra}, the generator-discriminator pre-training framework, has achieved impressive semantic construction capability among various downstream tasks. Despite the convincing performance, ELECTRA still faces the challenges of monotonous training and deficient interaction. Generator with only masked language modeling (MLM) leads to biased learning and label imbalance for discriminator, decreasing learning efficiency; no explicit feedback loop from discriminator to generator results in the chasm between these two components, underutilizing the course learning. In this study, a multi-perspective course learning (MCL) method is proposed to fetch a many degrees and visual angles for sample-efficient pre-training, and to fully leverage the relationship between generator and discriminator. Concretely, three self-supervision courses are designed to alleviate inherent flaws of MLM and balance the label in a multi-perspective way. Besides, two self-correction courses are proposed to bridge the chasm between the two encoders by creating a ``correction notebook'' for secondary-supervision. Moreover, a course soups trial is conducted to solve the ``tug-of-war'' dynamics problem of MCL, evolving a stronger pre-trained model. Experimental results show that our method significantly improves ELECTRA's average performance by 2.8\% and 3.2\% absolute points respectively on GLUE and SQuAD 2.0 benchmarks, and overshadows recent advanced ELECTRA-style models under the same settings. The pre-trained MCL model is available at \href{https://huggingface.co/McmanusChen/MCL-base}{https://huggingface.co/McmanusChen/MCL-base}.
\end{abstract}

\section{Introduction}

Language models pre-training~\cite{gpt,gpt2,bert,roberta} has shown great success in endowing machines with the ability to understand and process various downstream NLP tasks.
A wide range of pre-training strategies have been proposed, among which the most prevailing object is the masked language modeling (MLM)~\cite{bert}.
Such autoencoding language modeling objects~\cite{dae} typically first randomly corrupt a certain percentage of training corpus with masked tokens, and then encourage encoders to restore the original corpus.
To reduce the randomness of pre-training and produce a sample-efficient method, ELECTRA-style frameworks~\cite{electra} leverage an Transformer-based~\cite{transformer} generator training with MLM to build challenging ennoising sentences for a discriminator in the similar structure to carry out the denoising procedure. 

Typically in the ELECTRA-style training, the generator first constructs its semantic representations through MLM training and cloze these masked sentences with pseudo words; in the meanwhile, the discriminator inherits the information from the former and distinguish the originality of every token, which is like a step-by-step course learning.
However, only MLM-based generator training may lead to monotonous learning of data, which conduces to the incomprehensive generation and imbalanced label of corrupted sentences for the discriminator~\cite{hao2021learning}. Besides, interactivity between the two encoders stops abruptly except the sharing of embedding layers~\cite{mc-bert,coco-lm}, since there is no direct feedback loop from discriminator to generator.

To enhance the efficiency of training data and to adequately utilize the relationship of generator and discriminator, in this work, we propose a sample-efficient method named {\FULLMODELNAME{}} ({\MODELNAME{}}). In the first phase of {\MODELNAME{}}, to fetch a many degrees and visual angles to impel initial semantic construction, three self-supervision courses are designed, including cloze test, word rearrangement and slot detection. These courses instruct language models to deconstruct and dissect the exact same corpus from multi perspectives under the ELECTRA-style framework. In the second phase, two self-correction courses are tasked to refine both generator and discriminator. A confusion matrix regarding to discriminator's recognition of each sentence is analyzed and applied to the construction of revision corpora. Secondary learning is carried out for the two components in response to the deficiencies in the previous course learning. At last, the model mines the same batch of data from multiple perspectives, and implement progressive semantic learning through the self-correction courses. 

Experiments on the most widely accepted benchmarks GLUE~\cite{wang2018glue} and SQuAD 2.0~\cite{squad2} demonstrate the effectiveness of the proposed  {\MODELNAME{}}. Compared with previous advanced systems, {\MODELNAME{}} achieved a robust advantage across various downstream tasks. 
Abundant ablation studies confirm that multi-perspective courses encourage models to learn the data in a sample-efficient way.
Besides, a course soups trial is conducted to further interpret and dissect the core of multi-perspective learning, providing a novel approach to enhance the pre-training efficiency and performance.

\section{Preliminary} \label{sec: pre}

In this work, we built our system based on the ELECTRA-style framework.
Thus, the framework of ELECTRA is reviewed. 
Unlike BERT~\cite{bert}, which uses only one transformer encoder trained with MLM, ELECTRA is trained with two transformer encoders: a generator \(G\) and a discriminator \(D\). 
\(G\) is trained with MLM and used to generate ambiguous tokens to replace masked tokens in the input sequence.
Then the modified input sequence is fed to \(D\), which needs to determine if a corresponding token is either an original token or a token replaced by the generator.

\paragraph{Generator Training}
Formally, given an input sequence \(X = [x_1, x_2, ..., x_n]\), a mask operation is conducted to randomly replace its tokens with \(\texttt{[MASK]}\) at the position set $\bm{r}$.\footnote{Typically the proportion is set as 15\%, which means 15\% of the tokens are masked out for each sentence.}
And the masked sentence \(X^\text{mask} = [x_1, x_2, ..., \texttt{[MASK]}_i, ..., x_n]\) is fed into the generator to produce the contextualized representations $\{\bm{h}_i\}_{i=1}^n$. \(G\) is trained via the following loss $\mathcal{L}_{\text{MLM}}$ to predict the original tokens from the vocabulary \(V\) at the masked positions:
\begin{equation}
p_{\text{MLM}} (x_t | \bm{h}_i) = \frac{\exp(\bm{x}_t^\top \bm{h}_i)}{\sum_{t'=1}^{|V|} \exp(\bm{x}_{t'}^\top \bm{h}_i)} ,
\quad
\end{equation}
\begin{equation}
\mathcal{L}_{\text{MLM}} = \mathbb{E} \left( - \sum_{i\in \bm{r}} \log  p_{\text{MLM}} \left( x_i \big| \bm{h}_i \right) \right) ,
\end{equation}
where $\{\bm{x}_t\}_{t=1}^{|V|}$ are the embeddings of tokens that are replaced by $\texttt{[MASK]}$. Masked language modeling only conducts on the masked positions.

\paragraph{Discriminator Training.} 
\(G\) tends to predict the original identities of the masked-out tokens and thus $X^\text{rtd}$ is created by replacing the masked-out tokens with generator samples:
\begin{equation}
  x_i^\text{rtd} \sim p_{\text{MLM}} \left(x | \bm{h}_i \right),\, \text{if $i \in \bm{r}$ };
  \quad
  x_i^\text{rtd} = x_i,\, \text{else.}
\end{equation}

$D$ is trained to distinguish whether the tokens in $X^\text{rtd}$ have been replaced by $G$ via the replaced token detection (RTD) loss $\mathcal{L}_{\text{RTD}}$:
\begin{equation}
p_\text{RTD} (x_i^\text{rtd} = x_i \big| \bm{h}_i ) =  \frac{\exp(\bm{w}^\top \bm{h}_i)}{(1 + \exp(\bm{w}^\top \bm{h}_i))},
\end{equation}
\begin{equation}
\begin{split}
\mathcal{L}_{\text{RTD}} = \mathbb{E} \Bigg( - \sum_{x_i^\text{rtd} = x_i} \log p_{\text{RTD}}\left(x_i^\text{rtd} = x_i \big| \bm{h}_i \right) \\
- \sum_{x_i^\text{rtd} \neq x_i} \log \left(1 - p_{\text{RTD}}\left(x_i^\text{rtd} = x_i \big| \bm{h}_i \right) \right) \Bigg),
\end{split}
\end{equation}
where $\bm{w}$ is a learnable weight vector. This optimization is conducted on all tokens.

The overall pre-training objective is defined as:
\begin{equation}
\mathcal{L}_{\text{ELECTRA}} = \mathcal{L}_{\text{MLM}} + \lambda \mathcal{L}_{\text{RTD}} .
\end{equation}
where $\lambda$ (typically 50) is a hyperparameter used to balance the training pace of $G$ and $D$. Only $D$ is fine-tuned on downstream tasks after pre-training.

\section{Challenges} \label{sec:cha}

\paragraph{Biased Learning}
Though the ELECTRA training method is simple and effective, treating corpora from a single perspective could cause biased learning. 
As for the progress of MLM and RTD, there exists an inherent flaw that $G$ might predict appropriate but not original token on the $\texttt{[MASK]}$ position, and such appropriate expression still needs to be judged as substitution by $D$. For example, if the original sequence ``Alan buys an apple'' is masked as ``Alan $\texttt{[MASK]}$ an apple'', there are too many candidate words like ``eats, peels, cuts'' that can replace $\texttt{[MASK]}$ to form a harmonious context. Thus, to request $D$ to continue to distinguish replaced tokens is a difficult even awkward task. Treating the same piece of data in a single way reduces the effectiveness of training.
As for the distribution of generated corpus from $G$, the label-imbalance may gradually emerge with the MLM training of $G$, which could disturb the RTD training of $D$. As the semantic construction of $G$ thrives with the pre-training, the pseudo token on $\texttt{[MASK]}$ becomes more reasonable and even matches the original word. Thus, the proportion of replaced tokens in the training sentences of $D$ collapses, which interferes with the binary classification task seriously.
In this work, three self-supervision courses are tasked to train the model in a multi-perspective way, improving the learning efficiency of data and balancing the distribution of labels.

\paragraph{Deficient Interaction}
The core of self-supervision training is to ingeniously design and construct labeled data from original corpora. Evolving from random masking as BERT does, ELECTRA provides more realistic corpora by generator samples, encouraging $G$ and $D$ to compete with each other. However, there is no explicit feedback loop from $D$ to $G$, resulting that the pre-training of $G$ is practically dominated by MLM as before. To bridge the chasm between these two components, in this work, we take advantage of discriminant results from $D$ to create a ``correction notebook'' for both $G$ and $D$, and propose two self-correction courses to provide secondary-supervision.
Revision-training fully leverages the relationship and characteristics of the two encoders to increase the quality of training.

\section{Methodology}
In this section, we will start by formulating three self-supervision courses which encourage models to treat data in a multi-perspective way. Then two self-correction courses are elaborated, deriving from the course-like relationship between $G$ and $D$. These various courses are weaved into the entirety of the {\FULLMODELNAME} method.

\subsection{Self-supervision Course}
The essentiality of large-scale data pre-training, undoubtedly is to excogitate a way to take full advantage of the massive rude corpora. ELECTRA has provided an applicable paradigm for models to construct semantic representations through ennoising and denosing. Based on this framework, we extend the perspective that models look at sequences and propose three binary classification tasks in order to improve training efficiency, alleviate biased learning, and balance label distributions. 

\begin{figure}[t]
\centering
\includegraphics[width=0.48\textwidth]{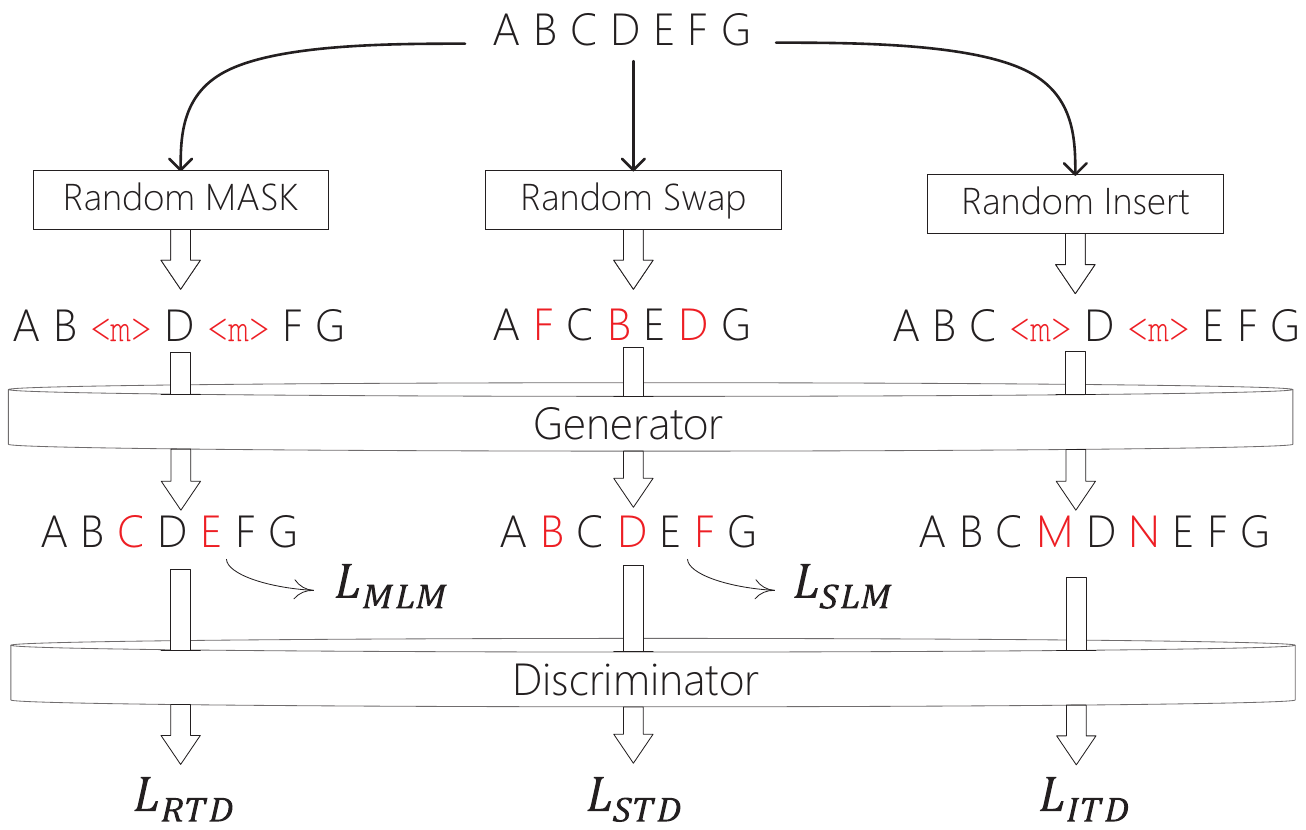}
\caption{The overall structure of the self-supervision courses. $\texttt{<m>}$ denotes $\texttt{[MASK]}$. A capital letter stands for a token and letters in red indicate operated positions.}
\label{fig:mcl}
\end{figure}

\subsubsection{Replaced Token Detection}
On account of the compelling performance of pre-training language models with masked language modeling, we retain the replaced token detection task from ELECTRA. Following the previous symbol settings, given an original input sequence \(X = [x_1, x_2, ..., x_n]\), we first mask out it into \(X^\text{mask} = [x_1, x_2, ..., \texttt{[MASK]}_i, ..., x_n]\), which is then fed into $G$ to get the filling-out sequence \(X^\text{rtd} = [x_1, x_2, ..., x^\text{rtd}_i, ..., x_n]\) by generator samples. Finally, $D$ is tasked to figure out which token is original or replaced.
As illustrated in the Section \ref{sec: pre}, $G$ and $D$ are trained with $\mathcal{L}_{\text{MLM}}$ and $\mathcal{L}_{\text{RTD}}$ respectively. MLM endows $G$ with fundamental contextual semantic construction by cloze test, and RTD is a higher level course for $D$ to let the model drill down into context for seeking out dissonance in the pseudo sequence $X^\text{rtd}$.

\subsubsection{Swapped Token Detection}
Intuitively, recombination tasks contribute to sequence-related learning. As mentioned in Section \ref{sec:cha}, information absence at the $\texttt{[MASK]}$ position will cause the unreliability of generated pseudo words. Whether the filled sample is appropriate or not, biased learning occurs interfering training of $D$. Thus, to reserve primeval message for precise prediction, without slashing the degree of task difficulty, we present swapped token detection (STD) course to sharpen the model's structure perception capability through a word rearrangement task.

For an input sentence \(X = [x_1, x_2, ..., x_n]\), a random position set $\bm{s}$ is chosen for swapping operation.\footnote{In the absence of special instructions, the proportion is set as 15\%.} 
Precisely, tokens at the chosen position are extracted, reordered and filled back into the sentence. $G$ is required to restore the swapped sentence \(X^\text{swap}\) to $X$, and the adjacent $D$ is tasked to discriminate which token is swapped in $X^\text{std}$ by generator samples. Note the contextualized representations from $G$ as $\{\bm{h}_i\}_{i=1}^n$, the training process of swapped language modeling (SLM) is formulated below:
\begin{equation}
p_{\text{SLM}} (x_s | \bm{h}_i) = \frac{\exp(\bm{x}_s^\top \bm{h}_i)}{\sum_{s'=1}^{|V|} \exp(\bm{x}_{s'}^\top \bm{h}_i)} ,
\quad
\end{equation}
\begin{equation}
\mathcal{L}_{\text{SLM}} = \mathbb{E} \left( - \sum_{i\in \bm{s}} \log  p_{\text{SLM}} \left( x_i \big| \bm{h}_i \right) \right) ,
\end{equation}
where $\{\bm{x}_s\}_{s=1}^{|V|}$ are the embeddings of tokens at the swapped positions. Note that the vocabulary \(V\) is still the same across all courses, because it helps the generation of $G$ in a consistent and natural environment, even the correct answer is lying in the pending sequence during SLM. SLM is only conducted on tokens at the swapped positions.

SLM brings $G$ to making reasonable even original predictions on the swapped positions, taking the attention of training from guessing of a single word to comprehensively understanding structure and logic of the whole sequence.
The swapped token detection (STD) course of $D$ is naturally formed as a deja vu binary classification. $X^\text{std}$ is created by replacing the swapped positions with generator samples:
\begin{equation}
  x_i^\text{std} \sim p_{\text{SLM}} \left(x | \bm{h}_i \right),\, \text{if $i \in \bm{s}$ };
  \quad
  x_i^\text{std} = x_i,\, \text{else.}
\end{equation}

$D$ is trained to distinguish whether the tokens in $X^\text{std}$ is original or not via the swapped token detection (RTD) loss:
\begin{equation}
p_\text{STD} (x_i^\text{std} = x_i \big| \bm{h}_i ) =  \text{sigmoid}(\bm{w}_s^T \bm{h}_i),
\end{equation}
\begin{equation}
\begin{split}
\mathcal{L}_{\text{STD}} = \mathbb{E} \Bigg( - \sum_{x_i^\text{std} = x_i} \log p_{\text{STD}}\left(x_i^\text{std} = x_i \big| \bm{h}_i \right) \\
- \sum_{x_i^\text{std} \neq x_i} \log \left(1 - p_{\text{STD}}\left(x_i^\text{std} = x_i \big| \bm{h}_i \right) \right) \Bigg).
\end{split}
\end{equation}
where $\bm{w}_s$ is an independent trainable parameter from $\bm{w}$ since each of courses uses its own binary classification head.

\subsubsection{Inserted Token Detection}
With the pace of pre-training with MLM and SLM, $G$ is inevitably able to produce much more harmonious sequences for the consummation of semantic learning. In the meanwhile, the label distribution of corrupted sentences provided by $G$ becomes magically imbalanced, since almost all tokens exactly match the words in the original sentence. Thus, training of $D$ faces serious interference and lack of efficiency. The propensity of the training labels leads to the propensity of $D$'s judgment.

To alleviate the issue of label-imbalance, and to seek another perspective of treating data, we propose the inserted token detection (ITD) course. For a given sentence \(X = [x_1, x_2, ..., x_n]\), $\texttt{[MASK]}$ is randomly inserted into the sequence at the inserted position set $\bm{i}$. The extended sentence \(X^\text{in}\) contains several illusory vacancies waiting for the prediction of $G$. Subsequently, $D$ has to figure out which token should not be presented in the generated sentence  \(X^\text{itd}\) with the training of the following loss:
\begin{equation}
p_\text{ITD} (x_i^\text{itd} = x_i^{in} \big| \bm{h}_i ) =  \text{sigmoid}(\bm{w}_i^T \bm{h}_i),
\end{equation}
\begin{equation}
\begin{split}
\mathcal{L}_{\text{ITD}} = \mathbb{E} \Bigg( - \sum_{x_i^\text{itd} = x_i^{in}} \log p_{\text{ITD}}\left(x_i^\text{itd} = x_i^{in} \big| \bm{h}_i \right) \\
- \sum_{x_i^\text{itd} \neq x_i^{in}} \log \left(1 - p_{\text{ITD}}\left(x_i^\text{itd} = x_i^{in} \big| \bm{h}_i \right) \right) \Bigg).
\end{split}
\end{equation}

On the one hand, the ratio of real and inserted words is fixed, solving the label-imbalance to some extent. On the other hand, training on void locations tones up the generation capability of models. 

The overall structure of the proposed self-supervision courses is presented in Figure~\ref{fig:mcl}. All courses are jointly conducted within the same data and computing steps.

\subsection{Self-correction Course}
According to the above self-supervision courses, a competition mechanism between $G$ and $D$ seems to shape up. Facing the same piece of data, $G$ tries to reform the sequence in many ways, while $D$ yearns to figure out 
all the jugglery caused previously. However, the shared embedding layer of these two encoders becomes the only bridge of communication, which is apparently insufficient. To strengthen the link between the two components, and to provide more supervisory information on pre-training, we conduct an intimate dissection of the relationship between $G$ and $D$.

\begin{table}[t]
\centering
\begin{tabular}{ccc}
\toprule
\multicolumn{1}{l}{Predict\textbackslash{}Label} & original   & replaced   \\
\midrule
\multirow{2}{*}{original}              & \checkmark    &  \ding{55}     \\
                                       & $pos_1$ & $pos_2$  \\\midrule
\multirow{2}{*}{replaced}              &  \ding{55}      & \checkmark  \\
                                       & $pos_3$ & $pos_4$ \\
                                       \bottomrule
\end{tabular}
\caption{The confusion matrix of output tokens from $D$. \checkmark denotes that $D$ makes a correct judgment, conversely \ding{55} presents the situation of wrong discrimination.}\label{tab:matrix}
\end{table}

Take the procedure of RTD for example. For each token $x_i^{rtd}$ in the corrupted sentence $X^\text{rtd}$, whereafter fed into $D$, we identify and document its label by comparing with the token $x_i$ at the corresponding position in $X$. After the discrimination process of $D$, this token is binary classified as original or replaced. As shown in Table~\ref{tab:matrix}, there exist four situations of distinguish results for $x_i$. $pos_1$: where $G$ predicts the correct answer on the $\texttt{[MASK]}$ position and $D$ successfully makes a good judgment, no additional operation needs to be conducted for this kind of token. $pos_2$: where $G$ fills an alternative to replace the original token and $D$ inaccurately views it as original, it means $G$ produces an appropriate expression to form a harmonious context as mentioned in Section~\ref{sec:cha}, which makes it difficult for $D$ to distinguish.
$pos_3$: where $D$ makes a clumsy mistake of incorrectly annotating an original token as replaced.
$pos_4$: where $G$ fills in an impertinent token at the $\texttt{[MASK]}$ position and $D$ easily figures it out.

\begin{figure}[t]
\centering
\includegraphics[width=0.48\textwidth]{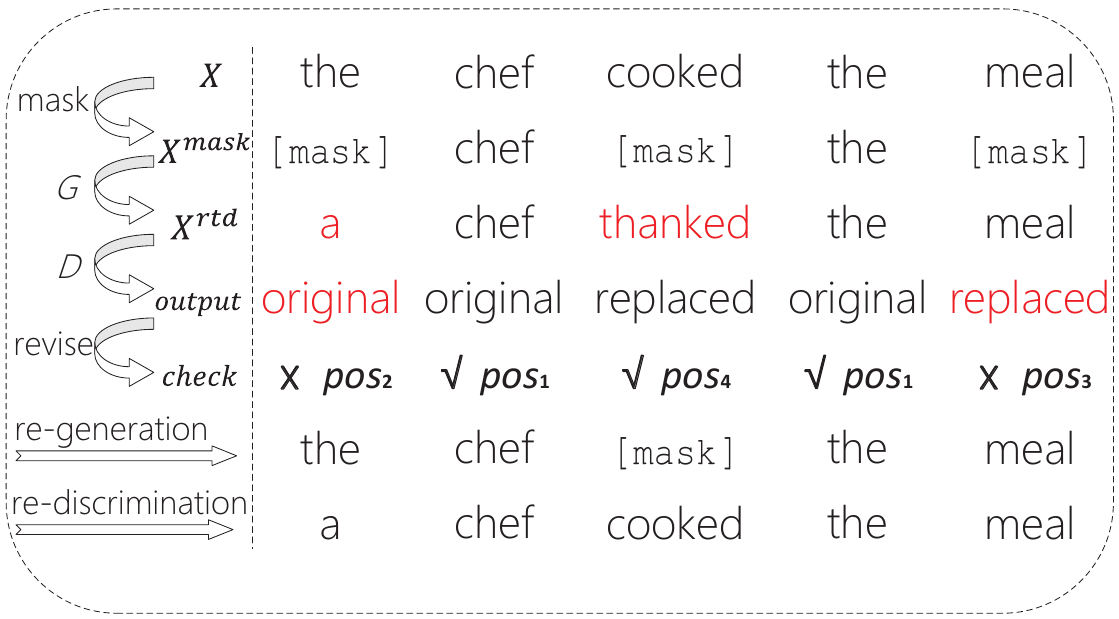}
\caption{An example for self-correction course of RTD.}  \label{fig:cobook}
\end{figure}

To sum it up, on the one hand, $G$ needs to re-generate tokens at $pos_4$, since the initial alternatives are inappropriate and unchallenging for $D$. 
As shown in Figure~\ref{fig:cobook}, too much $\texttt{[MASK]}$ are placed at important locations rich in information, leading to the erratic generation ``thanked''.
Considering that other $\texttt{[MASK]}$ in the same sequence may interfere with the generation of tokens at $pos_4$, we restore other $\texttt{[MASK]}$ to the original tokens for convenience of the re-generation process.
On the other hand, $D$ is expected to re-discriminate tokens at $pos_2$ and $pos_3$. When there exist tokens at $pos_4$ in a sequence, these inappropriate tokens may seriously disturb decisions of $D$ on other tokens, leading to the consequence of $pos_2$ and $pos_3$.
Take the sentence in Figure~\ref{fig:cobook} for example, serious distraction ``thanked'' makes $D$ falsely judges ``meal'' as replaced.
So we replace tokens at $pos_4$ in $X^\text{rtd}$ to original tokens to alleviate this kind of interference, and conduct the re-discrimination training on $D$ at $pos_2$ and $pos_3$.

By sorting out and analyzing errors, an ``correction notebook'' for $G$ and $D$ is built, guiding the re-generation and re-discrimination training. Note that it's not just redoing the problem, however, we redesign the context for each kind of issue. Thus, $\mathcal{L}_{\text{re-MLM}}$ and $\mathcal{L}_{\text{re-RTD}}$ is designed as the learning objective for self-correction course of RTD. Likewise, $\mathcal{L}_{\text{re-SLM}}$ and $\mathcal{L}_{\text{re-STD}}$ presents the training loss self-correction course of STD.\footnote{The equation is not listed since the form is consistent with the previous text.} 
Cause there are no original tokens at the inserted $\texttt{[MASK]}$ positions, no revision is conducted for ITD. Two proposed self-correction courses bridge the chasm between $G$ and $D$ through introspection and melioration, and provide a sample-efficient secondary-supervision for the same piece of data.

Finally, $G$ and $D$ are co-trained with three self-supervision courses as well as two self-correction courses. The proposed {\MODELNAME} dissects one same sequence profoundly and comprehensively, without incurring any additional inference or memory costs.

\begin{table*}[t]
\centering
\resizebox{\textwidth}{!}{
\begin{tabular}{llcccccccc}
\toprule
\multicolumn{1}{c}{\multirow{3}{*}{\textbf{Model}}}  & \multicolumn{9}{c}{\textbf{GLUE Single Task}} \\ \cmidrule(lr){2-10} 
\multicolumn{1}{c}{}                                               & \textbf{MNLI} & \textbf{QQP}           & \textbf{QNLI}          & \textbf{SST-2}         & \textbf{CoLA}          &\textbf{RTE}           & \textbf{MRPC}          & \textbf{STS-B}         & \textbf{AVG}                     \\
\multicolumn{1}{c}{}                       
& -m/-mm & Acc           & Acc          & Acc         & MCC          &Acc           & Acc          & PCC         &                      \\
\midrule
\multicolumn{10}{l}{\textit{Base} Setting: BERT   Base Size, Wikipedia + Book Corpus}                                                                                                                                                                                                        \\ \midrule
BERT~\cite{bert}                                                                               & 84.5/-           & 91.3          & 91.7          & 93.2          & 58.9          & 68.6          & 87.3          & 89.5          & 83.1                  \\
XLNet~\cite{xlnet}                                                                              & 85.8/85.4           & -         & -         & 92.7          & -         & -         & -         & -         & -             \\
RoBERTa~\cite{roberta}                                                                            & 85.8/85.5           & 91.3          & 92.0           & 93.7          & 60.1          & 68.2          & 87.3          & 88.5          & 83.3               \\
DeBERTa~\cite{deberta}                                                                            & 86.3/86.2           & -         & -         & -         & -         & -         & -         & -         & -            \\
TUPE~\cite{tupe}                                                                               & 86.2/86.2           & 91.3          & 92.2          & 93.3          & 63.6          & 73.6          & 89.9          & 89.2          & 84.9         \\
MC-BERT~\cite{mc-bert}                                                                            & 85.7/85.2           & 89.7          & 91.3          & 92.3          & 62.1          & 75.0           & 86.0           & 88.0           & 83.7              \\
ELECTRA~\cite{electra}                                                                 & 86.9/86.7           & 91.9          & 92.6          & 93.6          & 66.2          & 75.1          & 88.2          & 89.7          & 85.5               \\
+HP$_{\rm Loss}$+Focal~\cite{hao2021learning}                                                                & 87.0/86.9           & 91.7          & 92.7          & 92.6          & 66.7          & 81.3          & 90.7         & 91.0          & 86.7                 \\
CoCo-LM~\cite{coco-lm}                                                                            & \textbf{88.5}/88.3           & 92.0           & 93.1          & 93.2          & 63.9          & \textbf{84.8} & 91.4          & 90.3          & 87.2               \\
\textbf{{\MODELNAME}}                                                          & \textbf{88.5}/\textbf{88.5}  & \textbf{92.2} & \textbf{93.4} & \textbf{94.1} & \textbf{70.8} & 84.0          & \textbf{91.6} & \textbf{91.3} & \textbf{88.3} \\
\midrule
\multicolumn{10}{l}{\textit{Tiny} Setting: A quarter of training flops for ablation study, Wikipedia + Book Corpus}   \\
\midrule
ELECTRA(\emph{reimplement})                                                                  & 85.80/85.77          & 91.63          & 92.03          & 92.70          & 65.49          & 74.80          & 87.47          & 89.02          & 84.97                 \\
+STD                                                                              & 86.97/86.97          & 92.07          & 92.63          & 93.30          & 70.25          & 82.30          & 91.27          & 90.72          & 87.38                  \\
+ITD                                                                              & 87.37/87.33          & 91.87          & 92.53          & 93.40          & 68.45          & 81.37          & 90.87          & 90.52          & 87.08                  \\
Self-supervision                                                                    & 87.27/87.33          & 91.97          & 92.93          & 93.03          & 67.86          & 82.20          & 90.27          & 90.81          & 87.07                \\
+ re-RTD                                                                      & 87.57/87.50          & 92.07          & 92.67          & 92.97          & 69.80          & 83.27          & 91.60          & 90.71          & 87.57                   \\
+ re-STD                                                                       & 87.80/87.77          & 91.97          & 92.93          & 93.33          & \textbf{71.25} & 82.80 & 91.67 & \textbf{90.95} & \textbf{87.83}       \\
\textbf{{\MODELNAME}}                                                            & \textbf{87.90}/\textbf{87.83} & \textbf{92.13} & \textbf{93.00} & \textbf{93.47} & 68.81 & \textbf{83.03}          & \textbf{91.67} & 90.93 & 87.64  \\
\bottomrule
\end{tabular}}
\caption{All evaluation results on GLUE datasets for comparison. Acc, MCC, PCC denote accuracy, Matthews correlation, and Spearman correlation respectively. Reported results are medians over five random seeds.}\label{tab1}
\end{table*}

\section{Experiments}

\subsection{Setup}\label{sec:setup}

\paragraph{Pre-training Settings}
We implement the experiments on two settings: \textit{base} and \textit{tiny}.
\textit{Base} is the standard training configuration of BERT$_\text{Base}$~\cite{bert}.
The model is pre-trained on English Wikipedia and BookCorpus~\cite{bookcorpus}, containing 16 GB of text with 256 million samples.
We set the maximum length of the input sequence to 512, and the learning rates are 5e-4.
Training lasts 125K steps with a 2048 batch size. 
We use the same corpus as with CoCo-LM~\cite{coco-lm} and 64K cased SentencePiece vocabulary~\cite{kudo2018sentencepiece}. The details of the hyperparameter of pre-training is listed in Appendix~\ref{sec:appendix_pt}.
\textit{Tiny} conducts the ablation experiments on the same corpora with the same configuration as the \textit{base} setting, except that the batch size is 512.

\paragraph{Model Architecture}
The layout of our model architecture maintains the same as~\cite{coco-lm} both on \textit{base} and \textit{tiny} settings. $D$ consists of 12-layer Transformer, 768 hidden size, plus T5 relative position encoding ~\cite{t5}. $G$ is a shallow 4-layer Transformer with the same hidden size and position encoding. 
After pre-training, we discard $G$ and use $D$ in the same way as BERT, with a classification layer for downstream tasks.

\paragraph{Downstream Tasks}
To verify the effectiveness of the proposed methods, we conduct evaluation experiments on various downstream tasks.
We evaluate on the General Language Understanding Evaluation (GLUE) benchmark~\cite{wang2018glue} and Stanford Question Answering 2.0 (SQuAD 2.0) dataset~\cite{squad2}. 
As for the evaluation metrics of GLUE tasks, we adopt Spearman correlation for STS, Matthews correlation for CoLA, and accuracy for the other.
For SQuAD 2.0, in which some questions are unanswerable by the passage, the standard evaluation metrics of Exact-Match (EM) and F1 scores are adopted. 
More details of the GLUE and SQuAD 2.0 benchmark are listed in Appendix~\ref{sec:appendixA}.

\paragraph{Baselines}
Various pre-trained models are listed and compared in the \textit{base} setting. All numbers are from reported results in recent research. When multiple papers report different scores for the same method, we use the highest of them for comparison.

\paragraph{Implementation Details}
Our implementation builds upon the open-source implementation from fairseq~\cite{fairseq}. With 128 A100 (40 GB Memory), one pre-training run takes about 24 hours in \textit{base} setting. The fine-tuning costs are the same with BERT plus relative positive encodings. More details of fine-tuning are listed in Appendix~\ref{sec:appendix_ft}.

\begin{table}[t]
\centering
\resizebox{0.45\textwidth}{!}{
\begin{tabular}{lcc}
\toprule
\multicolumn{1}{c}{\multirow{2}{*}{\textbf{Model}}} & \multicolumn{2}{c}{\textbf{SQuAD 2.0}} \\ 
\cmidrule(lr){2-3} 
\multicolumn{1}{c}{}         & EM            & F1            \\ \midrule
\multicolumn{3}{l}{\textit{Base} Setting }                                                                                                                                                                                                        \\ \midrule
BERT~\cite{bert}                                  & 73.7          & 76.3          \\
XLNet~\cite{xlnet}                            & 78.5          & 81.3          \\
RoBERTa~\cite{roberta}                                & 77.7          & 80.5          \\
DeBERTa~\cite{deberta}                          & 79.3          & 82.5          \\
ELECTRA~\cite{electra}                      & 79.7          & 82.6          \\
+HP$_{\rm Loss}$+Focal~\cite{hao2021learning}                         & 82.7          & 85.4          \\
CoCo-LM~\cite{coco-lm}                            & 82.4          & 85.2          \\
\textbf{{\MODELNAME}}      &        \textbf{82.9} & \textbf{85.9} \\
\midrule
\multicolumn{3}{l}{\textit{Tiny} Setting for ablation study}   \\
\midrule
ELECTRA(\emph{reimplement})                                                                        & 79.37          & 81.31          \\
+STD                                                                                 & 81.73          & 84.55          \\
+ITD                                                                                 & 81.43          & 84.20          \\
Self-supervision                                                                        & 81.87          & 84.85          \\
+ re-RTD                                                                          & 81.70          & 84.48          \\
+ re-STD                                                                     & 81.81          & 84.71          \\
\textbf{{\MODELNAME}}                                                        & \textbf{82.04} & \textbf{84.93} \\
\bottomrule
\end{tabular}}
\caption{All evaluation results on SQuAD 2.0 datasets.}\label{tab2}
\end{table}

\subsection{Evaluation Results}
We first pre-trained our model with the proposed {\MODELNAME} method, and then fine-tuned it with training sets of 8 single tasks in the GLUE benchmark.
We conducted a hyperparameter search for all downstream tasks, and report the average scores among 5 random seeds.
Results are elaborated in the top half of Table~\ref{tab1}.
The proposed {\MODELNAME} evidently enhances ELECTRA and achieves at least 1.1\% absolute overall improvements against state-of-art pre-trained language models on the GLUE benchmark under the \textit{base} setting.
For the most widely reported task MNLI, our model achieves 88.5/88.6 points on the matched/mismatched (m/mm) set, which obtains 1.6/1.8 absolute improvements against ELECTRA.
Take a broader look at all GLUE single tasks, {\MODELNAME} overshadows all previous models, except RTE tasks, where CoCo-LM takes a narrow lead.

We also evaluated the proposed {\MODELNAME} on the SQuAD 2.0 datasets, which is an important reading comprehension dataset that requires the machine to extract the answer span given a document along with a question.
The results of Exact-Match (EM) and F1 score (F1) are displayed in the top half of Table~\ref{tab2}.
Consistently, our model significantly improves the ELECTRA baseline and achieves a banner score compared with other same-size models. Specifically, under the \textit{base} setting, the proposed {\MODELNAME} improves the absolute performance over ELECTRA by 3.2 points (EM) and 3.3 points (F1). Also, our model outperforms all other previous models with an overt margin.

The compelling results demonstrate the effectiveness of the proposed {\MODELNAME}.
With the equal amount of training corpus, plus slight computing cost of forward propagation, {\MODELNAME} tremendously advanced ELECTRA baseline, showing its property of sample-efficiency. In other words, multi-perspective course learning gives the model a deeper and more comprehensive insight into the underlying meaning of corpora, which provides more valuable information for the pre-training process.

\subsection{Ablation Study}
In order to dive into the role of each component in the proposed {\MODELNAME}, an ablation study is conducted under the \textit{tiny} setting. Both the GLUE and SQuAD 2.0 datasets are utilized for evaluation, and the ablation results are listed in the bottom half of Table~\ref{tab1} and Table~\ref{tab2}. Bolstered by several curve graphs regarding with loss and accuracy during pre-training, every course is discussed below. \footnote{Due to the constraints of space, curve graphs of pre-training and concrete instructions are replaced at Appendix~\ref{sec:appendix_curve}}

\paragraph{RTD}
The most basic component, also represents the ELCETRA itself.
Its performance would be employed as the baseline to compare with other additions. Not only the scores, but also the curves would be taken for important reference.

\paragraph{STD}
This course is tasked to help the model to obtain better structure perception capability through a more harmonious contextual understanding. STD improves ELECTRA on all tasks in GLUE and SQuAD 2.0 datasets. It is worth noting that the scores on CoLA task surprisingly stand out amongst the crowd.
The Corpus of Linguistic Acceptability (CoLA) is used to predict whether an English sentence is linguistically acceptable or not.
Apparently, pre-training on word rearrangement indeed lifts the global intellection of models, which makes it focus more on structure and logic rather than word prediction. Even the best CoLA result of 71.25 comes from the re-STD course, which further embodies the effectiveness of STD.

\paragraph{ITD}
This course is tasked to alleviate label-imbalance. As shown in Figure~\ref{fig:curves}, replace rate reflects the prediction accuracy of $G$. 
Accompanied by MLM and SLM, $G$ predicts more correct words on the $\texttt{[MASK]}$ positions, causing the ``replaced'' labels to become scarce for the training of $D$. By adding inserted $\texttt{[MASK]}$, the replace rate has a fixed lower limit corresponding to the inserted proportion, leading to a balanced distribution of labels. Besides, ITD shows great improvements over ELECTRA, especially on SST-2 datasets.
The Stanford Sentiment Treebank (SST-2) provides a dataset for sentiment classification that needs to determine whether the sentiment of a sentence extracted from movie reviews is positive or negative.
Predicting for the illusory $\texttt{[MASK]}$ makes the model focus more on content comprehension, which may helpful for sentiment classification.

\paragraph{Self-correction Course}
Revision always acts as a difficult assignment, because of the challenge to reverse stereotypes.
As shown in Figure~\ref{fig:curves}, losses of $G$ and $D$ during self-correction training generally exceed that during self-supervision training, demonstrating the difficulties.
However, the replace accuracy of re-RTD course goes higher than the baseline, certifying the effectiveness.
Despite that self-correction training outperforms other components on all downstream tasks, the phenomena of “tug-of-war” dynamics is worth exploring.
Scores listed in the last three rows of Table~\ref{tab1} almost touch each other, and optimal results of every single task do not always appear under the same model. It means multi-perspective courses may interfere with each other in attempts to pull parameters in different directions, which seems even more apparent under the self-correction course where secondary-samples are well designed for bootstrapping. To alleviate this situation and further improve the effectiveness of training, we found a feasible solution elaborated in Section~\ref{sec:soups}.

\begin{figure}[t]
\centering
\includegraphics[width=0.48\textwidth]{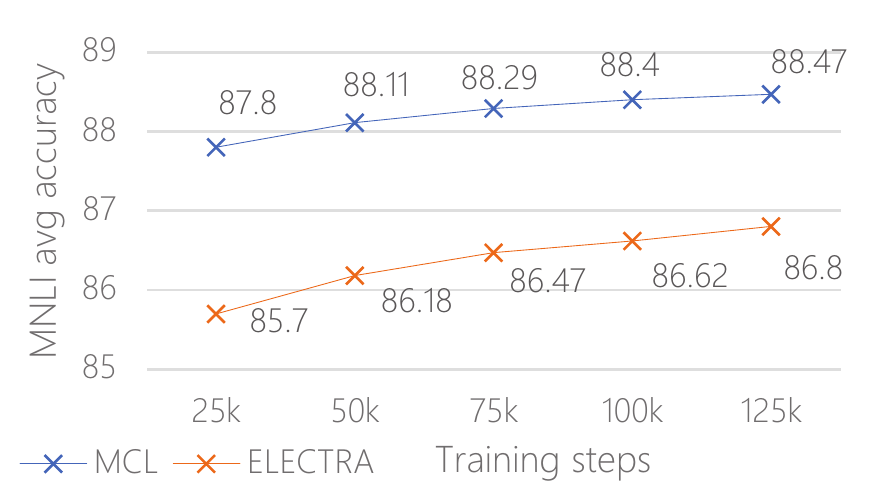}
\caption{MNLI Comparison of pre-training efficiency.}  \label{fig:steps}
\end{figure}

\subsection{Sample Efficiency Comparison}
To demonstrate the proposed {\MODELNAME} is sample-efficient, we conduct a comparative trial between {\MODELNAME} and ELECTRA. As shown in Figure~\ref{fig:steps}, the prevalent task MNLI is chosen for evaluation. For every 25K steps of pre-training, we reserved the model and fine-tuned it with the same configuration mentioned in Section~\ref{sec:setup}. Obviously, {\MODELNAME} preponderates over ELECTRA baseline on every training node, which obtains 87.8 points at 25K steps, demonstrating its enormous learning efficiency even on small pieces of corpora.

\begin{figure}[t]
\centering
\includegraphics[width=0.48\textwidth]{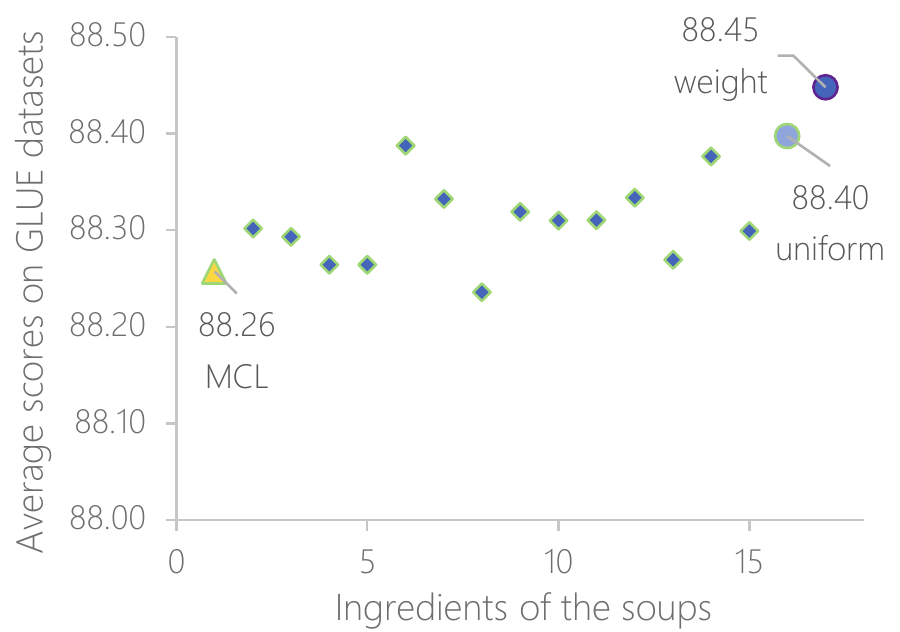}
\caption{Average GLUE results of the course soups.}  \label{fig:soups}
\end{figure}

\subsection{Course Soups Trial} \label{sec:soups}

Inspired by model soups \cite{modelsoup}, which averages many models in a hyperparameter sweep during fine-tuning, we find similarities and bring this idea to {\MODELNAME} in a task sweep during pre-training. Different courses lead the model to lie in a different low error basin, and co-training multiple courses may create the “tug-of-war” dynamics. To solve the training conflicts, and to further improve the learning efficiency of models in the later pre-training stage, we conduct a ``course soups'' trial.

For ingredients in soups, we arrange all combinations of 4 losses in self-correction courses, training them into 14 single models while retaining the structure of self-supervision courses. Then all ingredients are merged through uniform and weighted integration. 
Results lies in Figure~\ref{fig:soups}.
Optimal results obtained by weight soups, which improves the average GLUE score by 0.19 absolute points against our best model {\MODELNAME}. The results show that the course soups suggests a effective way to guide the later training of the model by separating multiple objectives and combining them at last. More details scores are listed in Appendix~\ref{sec:appendix_soups}.

\section{Conclusion}
This paper proposes the {\FULLMODELNAME} method, containing three self-supervision courses to improve learning efficiency and balance label distributions, as well as two self-correction courses to create a ``correction notebook'' for revision training.
Besides, the course soups method is designed to figure out a novel approach for efficient pre-training.
Experiments show that our method significantly improves ELECTRA's performance and overshadows multiple advanced models under same settings, verifying the effectiveness of {\MODELNAME}.

\section*{Limitations}
Although the proposed method has shown great performance to alleviate the issues of biased learning and deficient interaction, which are common problems among ELECTRA-style pre-training models, we should realize that the proposed method still can be further improved.
For example, the inherent flaw of RTD mentioned in Section~\ref{sec:cha} could only be relieved rather than solved. More about mission design regarding with this issue is worth studying.
Besides, although the results show great performance, more efforts are required to explore the hidden impact of each course, which will help the application of the proposed model in the future.


\appendix
\newpage

\section{Hyperparameters for Pre-training}
\label{sec:appendix_pt}

As shown in Table~\ref{tbl:pretraining_hyperparams}, we present the hyperparameters used for pre-training {\MODELNAME} on the \textit{base} setting.
We follow the optimization hyperparameters of CoCo-LM~\cite{coco-lm} for comparisons. Note that all losses conducted on $D$ are multiplied by $\lambda$ (set as 50), which is a hyperparameter used to balance the training pace of $G$ and $D$.

\begin{table}[h]
\centering
\begin{tabular}{lc}
\toprule
Layers & 12 \\
Hidden size & 768 \\
FFN inner hidden size & 3072 \\
Attention heads & 12 \\
Attention head size & 64 \\
Max relative position & 128 \\
Training steps & 125K \\
Batch size & 2048 \\
Adam $\epsilon$~\cite{adam} & 1e-6 \\
Adam $\beta$ & (0.9, 0.98) \\
Learning rate & 5e-4 \\
Learning rate schedule & Linear \\
Warmup steps & 10K\\
Gradient clipping & 2.0 \\
Dropout~\cite{hinton2012improving} & 0.1 \\
Weight decay & 0.01 \\
\bottomrule
\end{tabular}
\caption{
Hyperparameters for pre-training.
}
\label{tbl:pretraining_hyperparams}
\end{table}

\section{Details of Downstream Tasks}
\label{sec:appendixA}

GLUE contains a wide range of tasks covering textual entailment: RTE~\cite{giampiccolo2007third} and MNLI~\cite{mnli2017}, question-answer entailment: QNLI~\cite{squad1}, paraphrase: MRPC~\cite{mrpc2005}, question paraphrase: QQP~\cite{QQP}, textual similarity: STS~\cite{sts-b2017}, sentiment: SST~\cite{sst2013}, and linguistic acceptability CoLA~\cite{cola2018}. 

Natural Language Inference involves reading a pair of sentences and judging the relationship between their meanings, such as entailment, neutral and contradiction.
We evaluate on three diverse datasets, including Multi-Genre Natural Language Inference (MNLI), Question Natural Language Inference (QNLI) and Recognizing Textual Entailment (RTE).

Semantic similarity tasks aim to predict whether two sentences are semantically equivalent or not. The challenge lies in recognizing rephrasing of concepts, understanding negation, and handling syntactic ambiguity. Three datasets are used, including Microsoft Paraphrase corpus (MRPC), Quora Question Pairs (QQP)
dataset and Semantic Textual Similarity benchmark (STS-B).

Classification The Corpus of Linguistic Acceptability (CoLA) is used to predict whether an English sentence is linguistically acceptable or not. The Stanford Sentiment Treebank (SST-2) provides a dataset for sentiment classification that needs to determine whether the sentiment of a sentence extracted from movie reviews is positive or negative.

As a widely used MRC benchmark dataset, SQuAD 2.0 is a reading comprehension dataset that requires the machine to extract the answer span
given a document along with a question. We select the v2.0 version to keep the focus on the performance of pure span extraction performance. Two official metrics are used to evaluate the model performance: Exact Match (EM) and a softer metric F1 score, which measures the average overlap between the prediction and ground truth answer at the token level.

The summary of the General Language Understanding Evaluation (GLUE) benchmark~\cite{wang2018glue} is shown in Table~\ref{tbl:glue:datasets}.

\begin{table}[t]
\centering
\begin{tabular}{l l}
\toprule 
\textbf{Dataset} & \textbf{\#Train/\#Dev/\#Test}   \\ \midrule
\multicolumn{2}{l}{\emph{Single-Sentence Classification}} \\
CoLA (Acceptability)&8.5k/1k/1k \\
SST-2 (Sentiment)&67k/872/1.8k \\ \midrule
\multicolumn{2}{l}{\emph{Pairwise Text Classification}} \\
MNLI (NLI)& 393k/20k/20k\\
RTE (NLI) &2.5k/276/3k \\ 
QNLI (NLI)& 105k/5.5k/5.5k\\
WNLI (NLI) &634/71/146\\ 
QQP (Paraphrase)&364k/40k/391k\\ 
MRPC (Paraphrase) &3.7k/408/1.7k\\ \midrule
\multicolumn{2}{l}{\emph{Text Similarity}} \\
STS-B (Similarity) &7k/1.5k/1.4k \\ \bottomrule
\end{tabular}
\caption{Summary of the GLUE benchmark.
}
\label{tbl:glue:datasets}
\end{table}

\section{Hyperparameters for Fine-tuning}
\label{sec:appendix_ft}

Table~\ref{tbl:nlu_finetune_hyperparams} presents the hyperparameters used for fine-tuning over SQuAD v2.0~\cite{squad2}, and the GLUE benchmark~\cite{wang2018glue} following CoCo-LM for fair comparison.
On the development sets, the hyperparameters are searched based on the average performance of five runs.

\begin{table*}[t]
\centering
\resizebox{\textwidth}{!}{
\begin{tabular}{l*{3}{c}}
\toprule
Parameters & GLUE Small Tasks & GLUE Large Tasks & SQuAD 2.0  \\
\midrule
Max Epochs & \{2, 3, 5, 10\} & \{2, 3, 5\} & \{2, 3\} \\
Peak Learning Rate &  \{2e-5, 3e-5, 4e-5, 5e-5\} &  \{1e-5, 2e-5, 3e-5, 4e-5\} & \{2e-5, 3e-5, 4e-5, 5e-5\}\\
Batch Size & \{16, 32\} & 32  & \{16, 32\} \\
Learning Rate Decay & Linear & Linear & Linear\\
Warm-Up Proportion & \{6\%, 10\%\} & 6\% & \{6\%, 10\%\} \\
Sequence Length & 512 & 512 & 512 \\
Adam $\epsilon$ & 1e-6 & 1e-6 & 1e-6\\
Adam ($\beta_1$, $\beta_2$) & (0.9, 0.98) & (0.9, 0.98) & (0.9, 0.98)\\
Clip Norm & - & - & - \\
Dropout & 0.1 & 0.1  & 0.1\\
Weight Decay & 0.01 & 0.01 & 0.01 \\
\bottomrule
\end{tabular}}
\caption{Hyperparameter ranges searched for fine-tuning on GLUE and SQuAD 2.0. GLUE small tasks include {CoLA}, {RTE}, {MRPC} and {STS-B}. GLUE large tasks include {MNLI}, {QQP}, {QNLI} and {SST-2}.}
\label{tbl:nlu_finetune_hyperparams}
\end{table*}

\section{Curves for Ablation Study}
\label{sec:appendix_curve}

\begin{figure}[t]
\centering
\includegraphics[width=0.48\textwidth]{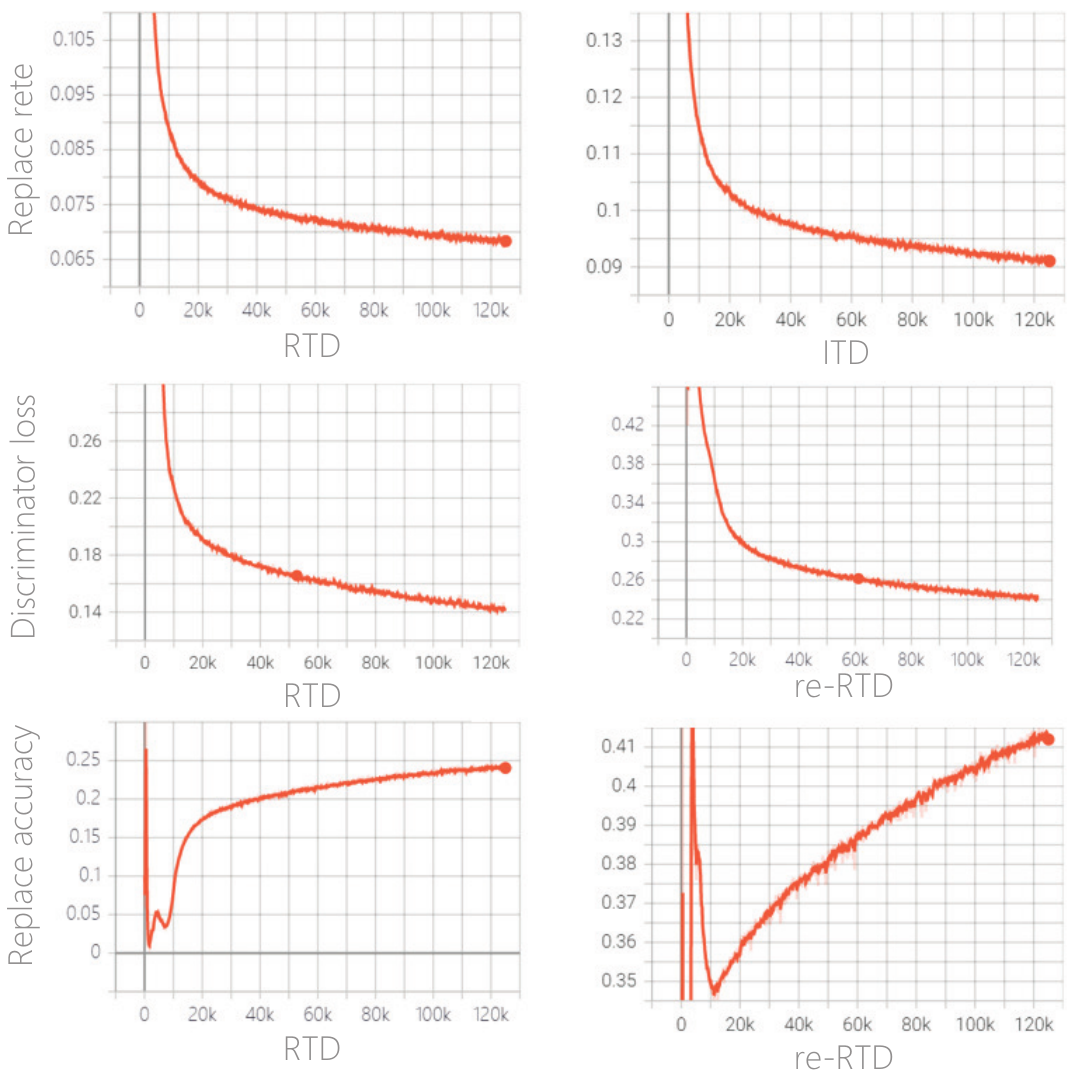}
\caption{Curves of replace rate, pre-training loss and replace accuracy for ablation study. All figures are drawn by Tensorboard.}  \label{fig:curves}
\end{figure}

As shown in Figure~\ref{fig:curves}, three metrics are selected to evaluate the quality of pre-training.
\paragraph{Replace Rate} This metric represents the ratio of replaced tokens in the corrupted sentence $X^\text{rtd}$. The less of this ratio, the better of $G$'s pre-training, and the more uneven of the label distributions for $D$. From the first row in the diagram, we can see that the lower bound of the replace rate with ITD apparently exceeds that with RTD, demonstrating that ITD indeed alleviates the issue of label-imbalance.

\paragraph{Loss and Replace Accuracy} Training loss reflects the pre-training assessment. One of the self-correction courses, re-RTD, holds a higher loss against that with RTD, showing the difficulty of the revision training. Replace accuracy denotes the prediction accuracy on replaced tokens. As shown in the third line of Figure~\ref{fig:curves}, re-RTD achieves better replace accuracy by a significant margin compared with RTD, demonstrating the effectiveness of the self-correction course.

\section{Detailed Scores of Course Soups}
\label{sec:appendix_soups}
Table~\ref{tab_soups} lists all scores on the GLUE benchmark of the course soups trial. It seems that optimal results on single tasks are randomly distributed in the ingredients of soups. Through uniform and weighted integration, the best model emerges with the uppermost average GLUE scores.

\begin{table*}[t]
\centering
\resizebox{\textwidth}{!}{
\begin{tabular}{llcccccccc}
\toprule
\multicolumn{1}{c}{\multirow{3}{*}{\textbf{Model}}}  & \multicolumn{9}{c}{\textbf{GLUE Single Task}} \\ \cmidrule(lr){2-10} 
\multicolumn{1}{c}{}                                               & \textbf{MNLI} & \textbf{QQP}           & \textbf{QNLI}          & \textbf{SST-2}         & \textbf{CoLA}          &\textbf{RTE}           & \textbf{MRPC}          & \textbf{STS-B}         & \textbf{AVG}                     \\
\multicolumn{1}{c}{}                       
& -m/-mm & Acc           & Acc          & Acc         & MCC          &Acc           & Acc          & PCC         &                      \\
\midrule
\multicolumn{10}{l}{\textit{Base} Setting: BERT   Base Size, Wikipedia + Book Corpus}                                                                                                                                                                                                        \\ \midrule
MCL        & 88.47/88.47          & 92.23          & 93.37          & 94.13          & 70.76          & 84.00          & 91.57          & 91.32          & 88.26          \\
$\mathcal{L}_{\text{re-MLM}}$  & \textbf{88.53}/88.50          & 92.23          & 93.40          & 94.33          & 70.53          & 84.00          & 92.00          & 91.18          & 88.30          \\
$\mathcal{L}_{\text{re-RTD}}$  & 88.47/88.43          & 92.23          & 93.37          & 94.13          & 70.77          & 83.63          & \textbf{92.40} & 91.20          & 88.29          \\
$\mathcal{L}_{\text{re-SLM}}$  & 88.43/88.43          & 92.23          & 93.43          & 94.27          & 70.53          & 83.77          & 92.00          & 91.29          & 88.26          \\
$\mathcal{L}_{\text{re-STD}}$  & 88.43/88.43          & 92.23          & 93.43          & 94.27          & 70.53          & 83.77          & 92.00          & 91.29          & 88.26          \\
$\mathcal{L}_{\text{re-MLM+RTD}}$ & 88.43/88.33          & 92.20          & 93.43          & 94.27          & 70.88          & 83.77          & 92.17          & 91.30          & 88.31          \\
$\mathcal{L}_{\text{re-MLM+SLM}}$ & 88.50/88.43          & 92.17          & 93.47          & 94.07          & 71.12          & 83.40          & \textbf{92.40} & 91.24          & 88.31          \\
$\mathcal{L}_{\text{re-MLM+STD}}$ & 88.43/88.43          & 92.23          & 93.37          & 94.27          & 71.09          & 83.77          & 92.17          & 91.24          & 88.33          \\
$\mathcal{L}_{\text{re-RTD+SLM}}$ & 88.47/88.43          & \textbf{92.27} & 93.43          & 94.23          & 70.84          & 83.27          & 92.23          & 91.25          & 88.27          \\
$\mathcal{L}_{\text{re-RTD+STD}}$ & 88.50/88.50          & 92.23          & 93.40          & 94.30          & 71.00          & 84.03          & 92.17          & 91.25          & 88.38          \\
$\mathcal{L}_{\text{re-SLM+STD}}$ & 88.50/88.47          & \textbf{92.27} & 93.57          & 94.13          & 70.53          & 83.63          & 92.23          & \textbf{91.36} & 88.30          \\
$\mathcal{L}_{\text{re-SLM+RTD+STD}}$ & 88.47/88.43          & \textbf{92.27} & 93.57          & 94.07          & \textbf{71.61} & 83.77          & 92.10          & 91.21          & 88.39          \\
$\mathcal{L}_{\text{re-MLM+SLM+STD}}$ & 88.47/\textbf{88.53} & \textbf{92.27} & 93.40          & 94.40          & 70.79          & 84.00          & 91.90          & 91.24          & 88.33          \\
$\mathcal{L}_{\text{re-MLM+RTD+STD}}$ & 88.50/88.47          & 92.23          & 93.43          & 94.07          & 70.80          & 83.53          & 91.93          & 91.15          & 88.24          \\
$\mathcal{L}_{\text{re-MLM+RTD+SLM}}$ & 88.43/88.40          & 92.23          & \textbf{93.60} & 94.30          & 70.75          & \textbf{84.13} & 91.77          & 91.25          & 88.32          \\
uniform soups    & \textbf{88.53}/88.43          & 92.23          & 93.43          & \textbf{94.53} & 71.40          & 83.53          & 92.23          & 91.24          & 88.40          \\
weight soups     & 88.47/88.43          & 92.20          & 93.57          & 94.13          & \textbf{71.61} & 84.00          & \textbf{92.40} & 91.22          & \textbf{88.45} \\
\bottomrule
\end{tabular}}
\caption{All evaluation results on GLUE datasets for the course soups trial. Acc, MCC, PCC denote accuracy, Matthews correlation, and Spearman correlation respectively. Reported results are medians over five random seeds.}\label{tab_soups}
\end{table*}

\end{document}